\documentclass[11pt,a4paper]{article}

\usepackage{times,latexsym}
\usepackage{url}
\usepackage[T1]{fontenc}
\usepackage{graphicx}
\usepackage{booktabs}
\usepackage{threeparttable}
\usepackage{multirow}
\usepackage{multicol}
\usepackage{makecell}
\usepackage{color, colortbl}
\usepackage{amsmath}
\usepackage{amssymb}
\usepackage{svg}
\usepackage[final]{EMNLP2022}
\usepackage{xspace,mfirstuc,tabulary}
\usepackage{enumitem}
\usepackage{tikz}
\usepackage{pifont}%
\usepackage{linguex}
\usepackage{subcaption}

\RequirePackage{etoolbox}
\RequirePackage{xcolor}
\definecolor{Gray}{gray}{0.9}
\definecolor{cb-black}      {RGB}{ 0,   0,   0}
\definecolor{cb-blue-green} {RGB}{ 0,  073,  073}
\definecolor{cb-green-sea}  {RGB}{ 0, 146, 146}
\definecolor{cb-rose}       {RGB}{255, 109, 182}
\definecolor{cb-salmon-pink}{RGB}{255, 182, 119}
\definecolor{cb-purple}     {RGB}{ 73,   0, 146}
\definecolor{cb-blue}       {RGB}{ 0, 109, 219}
\definecolor{cb-lilac}      {RGB}{182, 109, 255}
\definecolor{cb-blue-sky}   {RGB}{109, 182, 255}
\definecolor{cb-blue-light} {RGB}{182, 219, 255}
\definecolor{cb-burgundy}   {RGB}{146,   0,   0}
\definecolor{cb-brown}      {RGB}{146,  73,   0}
\definecolor{cb-clay}       {RGB}{219, 209,   0}
\definecolor{cb-green-lime} {RGB}{ 36, 255,  36}
\definecolor{cb-yellow}     {RGB}{255, 255, 109}

\usepackage{stfloats}

\newcommand{\rucola}[0]{\raisebox{-.25\height}{\includegraphics[width=.05\textwidth]{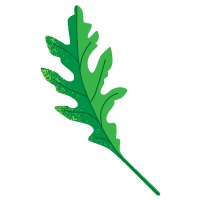}}}

\newcommand{\cmark}{\checkmark}%
\newcommand{\xmark}{$\times$}%

\makeatletter
\newcommand*{\radiobutton}{%
  \@ifstar{\@radiobutton0}{\@radiobutton1}%
}
\newcommand*{\@radiobutton}[1]{%
  \begin{tikzpicture}
    \pgfmathsetlengthmacro\radius{height("X")/3}
    \draw[radius=\radius] circle;
    \ifcase#1 \fill[radius=.6*\radius] circle;\fi
  \end{tikzpicture}%
}
\makeatother

\let\oldpm\pm
\renewcommand{\pm}{$\oldpm$}

\usepackage{todonotes}

\title{\rucola RuCoLA: Russian Corpus of Linguistic Acceptability}

\author{
    ~\textbf{Vladislav Mikhailov\textsuperscript{1}}\thanks{\ \ Equal contribution.}, 
    ~\textbf{Tatiana Shamardina\textsuperscript{2}$^*$}, 
    ~\textbf{Max Ryabinin\textsuperscript{3,4}$^*$} \\
    ~\textbf{Alena Pestova\textsuperscript{3}}, 
    ~\textbf{Ivan Smurov\textsuperscript{2}}, 
    ~\textbf{Ekaterina Artemova\textsuperscript{5,6}} \\
    \textsuperscript{1}SberDevices,
    \textsuperscript{2}ABBYY, 
    \textsuperscript{3}HSE University, \\ 
    \textsuperscript{4}Yandex,
    \textsuperscript{5}Huawei Noah's Ark Lab,  \\
    \textsuperscript{6}Center for Information and Language Processing (CIS), MaiNLP lab, LMU Munich, Germany \\
    \small{
    \textbf{Correspondence:} \href{mailto:vmikhailovhse@gmail.com}{\texttt{vmikhailovhse@gmail.com}}
}}

\begin{document}

\maketitle

\begin{abstract}
Linguistic acceptability (LA) attracts the attention of the research community due to its many uses, such as testing the grammatical knowledge of language models and filtering implausible texts with acceptability classifiers.
However, the application scope of LA in languages other than English is limited due to the lack of high-quality resources.
To this end, we introduce the Russian Corpus of Linguistic Acceptability (RuCoLA), built from the ground up under the well-established binary LA approach. 
RuCoLA consists of $9.8$k in-domain sentences from linguistic publications and $3.6$k out-of-domain sentences produced by generative models. The out-of-domain set is created to facilitate the practical use of acceptability for improving language generation.
Our paper describes the data collection protocol and presents a fine-grained analysis of acceptability classification experiments with a range of baseline approaches.
In particular, we demonstrate that the most widely used language models still fall behind humans by a large margin, especially when detecting morphological and semantic errors. We release RuCoLA, the code of experiments, and a public leaderboard\footnote{Available at~\href{https://rucola-benchmark.com/}{\texttt{rucola-benchmark.com}}} to assess the linguistic competence of language models for Russian.
\end{abstract}

\section{Introduction}
Recent NLP research has approached the linguistic competence of language models (LMs) with \emph{acceptability judgments}, which reflect a sentence's well-formedness and naturalness from the perspective of native speakers~\cite{chomsky1965aspects}. These judgments have formed an empirical foundation in generative linguistics for evaluating humans' grammatical knowledge and language acquisition~\cite{schutze1996empirical,sprouse2018acceptability}.

Borrowing conventions from linguistic theory, the community has put much effort into creating linguistic acceptability (LA) resources to explore whether LMs acquire grammatical concepts pivotal to human linguistic competence~\cite{kann-etal-2019-verb,warstadt-etal-2019-neural,warstadt-etal-2020-blimp-benchmark}. Lately, similar non-English resources have been proposed to address this question in typologically diverse languages~\cite{trotta-etal-2021-monolingual-cross,volodina-etal-2021-dalaj,hartmann-etal-2021-multilingual,xiang-etal-2021-climp}. However, the ability of LMs to perform acceptability judgments in Russian remains understudied.

\begin{table}[t!]
\centering
\scriptsize
\resizebox{0.87\columnwidth}{!}{
\begin{tabular}{llrrl}
\toprule
 &\textbf{Language} & \textbf{Size}&\textbf{\%} \\

\midrule

\textbf{CoLA} & \textbf{English} & 10.6k & 70.5\\
\textbf{ItaCoLA} & \textbf{Italian} & 9.7k & 85.4\\
\midrule
\textbf{RuCoLA} & \textbf{Russian} & 13.4k & 71.8\\
\bottomrule 
\end{tabular}
}
\caption{Comparison of RuCoLA with related binary acceptability classification benchmarks: CoLA \cite{warstadt-etal-2019-neural} and ItaCoLA \cite{trotta-etal-2021-monolingual-cross}. \textbf{\%}=Percentage of acceptable sentences.
}\label{tab:comparison}
\vspace{-2ex}
\end{table}

To this end, we introduce the Russian Corpus of Linguistic Acceptability (RuCoLA), a novel benchmark of $13.4$k sentences labeled as acceptable or not. In contrast to related binary acceptability classification benchmarks  in~\autoref{tab:comparison}, RuCoLA combines in-domain sentences manually collected from linguistic literature and out-of-domain sentences produced by nine machine translation and paraphrase generation models. The motivation behind the out-of-domain set is to facilitate the practical use of acceptability judgments for improving language generation~\cite{kane-etal-2020-nubia,batra-etal-2021-building}. Furthermore, each unacceptable sentence is additionally labeled with four standard and machine-specific coarse-grained categories: morphology, syntax, semantics, and hallucinations~\cite{raunak-etal-2021-curious}.

The main contributions of this paper are the following: (i) We create RuCoLA, the first large-scale acceptability classification resource in Russian. (ii) We present a detailed analysis of acceptability classification experiments with a broad range of baselines, including monolingual and cross-lingual Transformer~\cite{vaswani2017attention} LMs, statistical approaches, acceptability measures from pretrained LMs, and human judgements. (iii) We release RuCoLA, the code of experiments\footnote{Both RuCoLA and the code of our experiments are available at \href{https://github.com/RussianNLP/RuCoLA}{\texttt{github.com/RussianNLP/RuCoLA}}}, and a leaderboard to test the linguistic competence of modern and upcoming LMs for the Russian language.

\section{Related work}
\subsection{Acceptability Judgments}
\paragraph{Acceptability Datasets}
The design of existing LA datasets is based on standard practices in linguistics~\cite{myers2017acceptability,sep-linguistics}: binary acceptability classification~\cite{warstadt-etal-2019-neural,kann-etal-2019-verb}, magnitude estimation~\cite{vazquez-martinez-2021-acceptability}, gradient judgments~\cite{lau2017grammaticality,sprouse2018colorless}, Likert scale scoring~\cite{brunato2020accompl}, and a forced choice between minimal pairs~\cite{marvin-linzen-2018-targeted,warstadt-etal-2020-blimp-benchmark}. Recent studies have extended the research to languages other than English: Italian~\cite{trotta-etal-2021-monolingual-cross}, Swedish~\cite{volodina-etal-2021-dalaj}, French~\cite{feldhausen2020testing}, Chinese~\cite{xiang-etal-2021-climp}, Bulgarian and German~\cite{hartmann-etal-2021-multilingual}. Following the motivation and methodology by~\citet{warstadt-etal-2019-neural}, this paper focuses on the binary acceptability classification approach for the Russian language.

\paragraph{Applications of Acceptability} Acceptability judgments have been broadly applied in NLP. In particular, they are used to test LMs' robustness~\cite{yin-etal-2020-robustness} and probe their acquisition of grammatical phenomena~\cite{warstadt2019linguistic,choshen-etal-2022-grammar,zhang-etal-2021-need}. LA has also stimulated the development of acceptability measures based on pseudo-perplexity~\cite{lau-etal-2020-furiously}, which correlate well with human judgments~\cite{lau2017grammaticality} and show benefits in scoring generated hypotheses in downstream tasks~\cite{salazar-etal-2020-masked}. Another application includes evaluating the grammatical and semantic correctness in language generation~\cite{kane-etal-2020-nubia,harkous-etal-2020-text,bakshi-etal-2021-structure,batra-etal-2021-building}.

\subsection{Evaluation of Text Generation}
Machine translation (or MT) is one of the first sub-fields which has established diagnostic evaluation of neural models~\cite{dong2021survey}. Diagnostic datasets can be constructed by automatic generation of contrastive pairs~\cite{burlot-yvon-2017-evaluating}, crowdsourcing annotations of generated sentences~\cite{lau2014measuring}, and native speaker data~\cite{anastasopoulos-2019-analysis}. Various phenomena have been analyzed, to name a few: morphology~\cite{burlot-etal-2018-wmt18}, syntactic properties~\cite{sennrich-2017-grammatical,wei-etal-2018-evaluating}, commonsense~\cite{he-etal-2020-box}, anaphoric pronouns~\cite{guillou-etal-2018-pronoun}, and cohesion~\cite{bawden-etal-2018-evaluating}.

Recent research has shifted towards overcoming limitations in language generation, such as copying inputs~\cite{liu-etal-2021-copying}, distorting facts~\cite{santhanam2021rome}, and generating hallucinated content~\cite{zhou-etal-2021-detecting}. \citet{maynez-etal-2020-faithfulness} and \citet{liu-etal-2022-token} propose datasets on hallucination detection. \textsc{ScareCrow}~\cite{dou-etal-2022-gpt} and \textsc{TGEA}~\cite{he-etal-2021-tgea} focus on taxonomies of text generation errors. Drawing inspiration from these works, we create the machine-generated out-of-domain set to foster text generation evaluation with acceptability.

\section{RuCoLA}
\begin{table}[t!]
\centering
\small
\setlength{\tabcolsep}{2pt}
\begin{tabular}{lrrl}

\toprule
\textbf{Source}&\textbf{Size}&\textbf{\%}
&\textbf{Content}\\

\midrule

\texttt{rusgram} & 563 & 49.7 & Corpus grammar \\
\citet{Testelets:2001} & 1335 & 73.9 & General syntax \\
\citet{Lutikova:2010} & 193 & 75.6 & Syntactic structures \\
\citet{mitrenina2017} & 54 & 57.4 & Generative grammar \\
\citet{Paducheva:2010} & 1308 & 84.3 & Semantics of tense \\
\citet{Paducheva:2004}  & 1374 & 90.8 & Lexical semantics \\
\citet{Paducheva:2013} & 1462 & 89.5 & Aspects of negation \\
\citet{Seliverstova:2004} & 2104 & 80.8 & Semantics \\
\citet{shavrina-etal-2020-humans} & 1444 & 36.6 & Grammar exam tasks \\

\midrule

\textbf{In-domain}&\textbf{9837}&\textbf{74.5}&\\

\midrule
Machine Translation & 1286 & 72.8 & English translations
\\
Paraphrase Generation & 2322 & 59.9 & Automatic paraphrases 
\\
\midrule
\textbf{Out-of-domain}&\textbf{3608}&\textbf{64.6}&\\

\midrule
\textbf{Total}&\textbf{13445}&\textbf{71.8}&\\
\bottomrule 
\end{tabular}

\caption{RuCoLA statistics by source. The number of in-domain sentences is similar to that of CoLA and ItaCoLA. \textbf{\%}=Percentage of acceptable sentences.}
\label{tab:sources}
\end{table}    

\begin{table*}[ht!]
\centering
\small
\resizebox{\textwidth}{!}{
\begin{tabular}
{clcll}
\toprule
\textbf{Label} & \textbf{Set} & \textbf{Category} & \textbf{Sentence}&\textbf{Source}
\\\midrule
\multirow{2}{*}{\cmark} & \multirow{2}{*}{\textbf{In-domain}} & \multirow{2}{*}{\xmark} &
    \textit{Ya obnaruzhil ego lezhaschego odnogo na krovati.}
& \multirow{2}{*}{\citet{Testelets:2001}}
\\
&&&I found him lying in the bed alone.&\\

\midrule
\multirow{2}{*}{\textbf{*}} & \multirow{2}{*}{\textbf{In-domain}} & \multirow{2}{*}{\textsc{Syntax}} &
\textit{Ivan prileg, \textcolor{cb-burgundy}{chtoby on otdokhnul}.}
& \multirow{2}{*}{\citet{Testelets:2001}} \\
&&&Ivan laid down \textcolor{cb-burgundy}{in order that he has a rest}.&
\\ \midrule
\multirow{2}{*}{\cmark} & \multirow{2}{*}{\textbf{Out-of-domain}} & \multirow{2}{*}{\xmark} &
\textit{Ja ne chital ni odnogo iz ego romanov.}
 & \multirow{2}{*}{\citet{artetxe-schwenk-2019-massively}}
\\
&&&I have not read any of his novels.& \\

\midrule
\multirow{2}{*}{\textbf{*}}
& \multirow{2}{*}{\textbf{Out-of-domain}} & \multirow{2}{*}{\textsc{Hallucination}} &

\textit{Ljuk ostanavlivaet udachu ot etogo.} 
 & \multirow{2}{*}{\citet{schwenk-etal-2021-wikimatrix}}
\\
&&& Luke stops luck from doing this.&\\

\bottomrule
\end{tabular}
}
\caption{A sample of RuCoLA. \textbf{*}=Unacceptable sentences. \cmark=Acceptable sentences. The examples are translated for illustration purposes.} \label{rucola_sample}
\vspace{-1ex}
\end{table*}

\subsection{Design} 
\label{section:sources}
RuCoLA consists of in-domain and out-of-domain subsets, as outlined in~\autoref{tab:sources}. Below, we describe the data collection procedures for each subset.

\paragraph{In-domain Set}
Here, the data collection method is analogous to CoLA. The in-domain sentences and the corresponding authors' acceptability judgments\footnote{We keep unacceptable sentences marked with the ``*'', ``*?'' and ``??'' labels.} are drawn from fundamental linguistic textbooks, academic publications, and methodological materials\footnote{The choice is also based on the ease of manual example collection, e.g., high digital quality of the sources and no need for manual transcription.}. The works are focused on various linguistic phenomena, including but not limited to general syntax~\cite{Testelets:2001}, the syntactic structure of noun phrases~\cite{Lutikova:2010}, negation~\cite{Paducheva:2013}, predicate ellipsis, and subordinate clauses (\texttt{rusgram}\footnote{A collection of materials written by linguists for a corpus-based description of Russian grammar. Available at: \href{http://rusgram.ru/}{\texttt{rusgram.ru}}  }). \citet{shavrina-etal-2020-humans} introduce a dataset on the Unified State Exam in the Russian language, which serves as school finals and university entry examinations in Russia. The dataset includes standardized tests on high school curriculum topics made by methodologists. We extract sentences from the tasks on Russian grammar, which require identifying incorrect word derivation and syntactic violations.

\paragraph{Out-of-domain Set} The out-of-domain sentences are produced by nine open-source MT and paraphrase generation models using subsets of four datasets from different domains: Tatoeba~\cite{artetxe-schwenk-2019-massively}, WikiMatrix~\cite{schwenk-etal-2021-wikimatrix}, TED~\cite{qi-etal-2018-pre}, and Yandex Parallel Corpus~\cite{antonova-misyurev-2011-building}. We use cross-lingual MT models released as a part of the \texttt{EasyNMT} library\footnote{\href{https://github.com/UKPLab/EasyNMT}{\texttt{github.com/UKPLab/EasyNMT}}}: OPUS-MT~\cite{tiedemann-thottingal-2020-opus}, M-BART50~\cite{tang2020multilingual} and M2M-100~\cite{fan2020englishcentric} of 418M and 1.2B parameters. Russian WikiMatrix sentences are paraphrased via the \texttt{russian-paraphrasers} library~\cite{fenogenova-2021-russian} with the following models and nucleus sampling strategy: ruGPT2-Large\footnote{\href{https://huggingface.co/sberbank-ai/rugpt2large}{\texttt{hf.co/sberbank-ai/rugpt2large}}} (760M), ruT5 (244M)\footnote{\href{http://huggingface.co/cointegrated/rut5-base-paraphraser}{\texttt{hf.co/cointegrated/rut5-base-paraphraser}}}, and mT5~\cite{xue-etal-2021-mt5} of Small (300M), Base (580M) and Large (1.2B) versions. The annotation procedure of the generated sentences is documented in \S\ref{annotation_machine}.

\subsection{Violation Categories}
\label{phenomenon_tax}
Each unacceptable sentence is additionally labeled with one of the four violation categories: morphology, syntax, semantics, and hallucinations. The annotation for the in-domain set is obtained through manual working with the sources. The categories are manually defined based on the interpretation of examples provided by the experts, topics covered by chapters, and the general content of a linguistic source. The out-of-domain sentences are annotated as described in \S\ref{annotation_machine}. 

\paragraph{Phenomena} The phenomena covered by RuCoLA are well represented in Russian theoretical and corpus linguistics and peculiar to modern generative models. We briefly summarize our informal categorization and list examples of the phenomena below:

\begin{enumerate}[leftmargin=1.5em,noitemsep,topsep=0.5pt]
    \item \textsc{Syntax}: agreement violations, corruption of word order, misconstruction of syntactic clauses and  phrases, incorrect use of appositions, violations of verb transitivity or argument structure, ellipsis, missing grammatical constituencies or words.

    \item \textsc{Morphology}: incorrect derivation or word building, non-existent words.
    
    \item \textsc{Semantics}: incorrect use of negation, violation of the verb’s semantic argument structure.
    
    \item \textsc{Hallucination}: text degeneration, nonsensical sentences, irrelevant repetitions, decoding confusions, incomplete translations, hallucinated content.
\end{enumerate}

\autoref{rucola_sample} provides a sample of several RuCoLA sentences, and examples for each violation category can be found in~\autoref{appendix:examples}.

\subsection{Annotation of Machine-Generated Sentences}
\label{annotation_machine}
The machine-generated sentences undergo a two-stage annotation procedure on Toloka~\cite{toloka}, a crowdsourcing platform for data labeling\footnote{\href{https://toloka.ai/}{\texttt{toloka.ai}}}. Each stage includes an unpaid training phase with explanations, control tasks for tracking annotation quality\footnote{Control tasks are used on Toloka as common practice for discarding results from bots or workers whose quality on these tasks is unsatisfactory. In our annotation projects, the tasks are manually selected or annotated by a few authors: about 200 and 500 sentences for Stages 1 and 2, respectively.}, and the main annotation task. Before starting, the worker is given detailed instructions describing the task, explaining the labels, and showing plenty of examples. The instruction is available at any time during both the training and main annotation phases. To get access to the main phase, the worker should first complete the training phase by labeling more than 70\% of its examples correctly~\cite{nangia-bowman-2019-human}. Each trained worker receives a page with five sentences, one of which is a control one. 

We collect the majority vote labels via a dynamic overlap\footnote{\href{https://toloka.ai/docs/guide/concepts/dynamic-overlap.html}{\texttt{toloka.ai/docs/dynamic-overlap}}} from three to five workers after filtering them by response time and performance on control tasks. Appendix~\ref{app:design_details} contains a detailed description of the annotation protocol, including response statistics and the agreement rates.

\paragraph{Stage 1: Acceptability Judgments}\label{paragraph:stage_one} The first annotation stage defines whether a given sentence is acceptable or not. Access to the project is granted to workers certified as native speakers of Russian by Toloka and ranked top-$60$\% workers according to the Toloka rating system. Each worker answers 30 examples in the training phase. Each training example is accompanied by an explanation that appears in an incorrect answer. The main annotation phase counts 3.6k machine-generated sentences. The pay rate is on average \$2.55/hr, which is twice the amount of the hourly minimum wage in Russia. Each of 1.3k trained workers get paid, but we keep votes from only 960 workers whose annotation quality rate on the control sentences is more than 50\%. We provide a shortened translated instruction and an example of the web interface in~\autoref{tab:binary} (see Appendix~\ref{appendix:instructions}). 

\paragraph{Stage 2: Violation Categories}\label{paragraph:stage_two}
The second stage includes validation and annotation of sentences labeled unacceptable on \textbf{Stage 1} according to five answer options: ``Morphology'', ``Syntax'', ``Semantics'', ``Hallucinations'' and ``Other''. The task is framed as a multi-label classification, i.e., the sentence may contain more than one violation in some rare cases or be re-labeled as acceptable. We create a team of 30 annotators who are undergraduate BA and MA in philology and linguistics from several Russian universities. The students are asked to study the works on CoLA~\cite{warstadt-etal-2019-neural}, \textsc{TGEA}~\cite{he-etal-2021-tgea}, and hallucinations~\cite{zhou-etal-2021-detecting}. We also hold an online seminar to discuss the works and clarify the task specifics. Each student undergoes platform-based training on 15 examples before moving onto the main phase of 1.3k sentences. The students are paid on average \$5.42/hr and are eligible to get credits for an academic course or an internship. Similar to one of the data collection protocols by \citet{parrish-etal-2021-putting-linguist}, this stage provides direct interaction between authors and students in a group chat. We keep submissions with more than 30 seconds of response time per page and collect the majority vote labels for each answer independently. Sentences having more than one violation category or labeled as ``Other'' by the majority are filtered out. The shortened instruction is presented in~\autoref{table:multilabel} (see Appendix~\ref{appendix:instructions}).

\begin{figure*}[ht!]
  \centering
  \includegraphics[width=\linewidth]{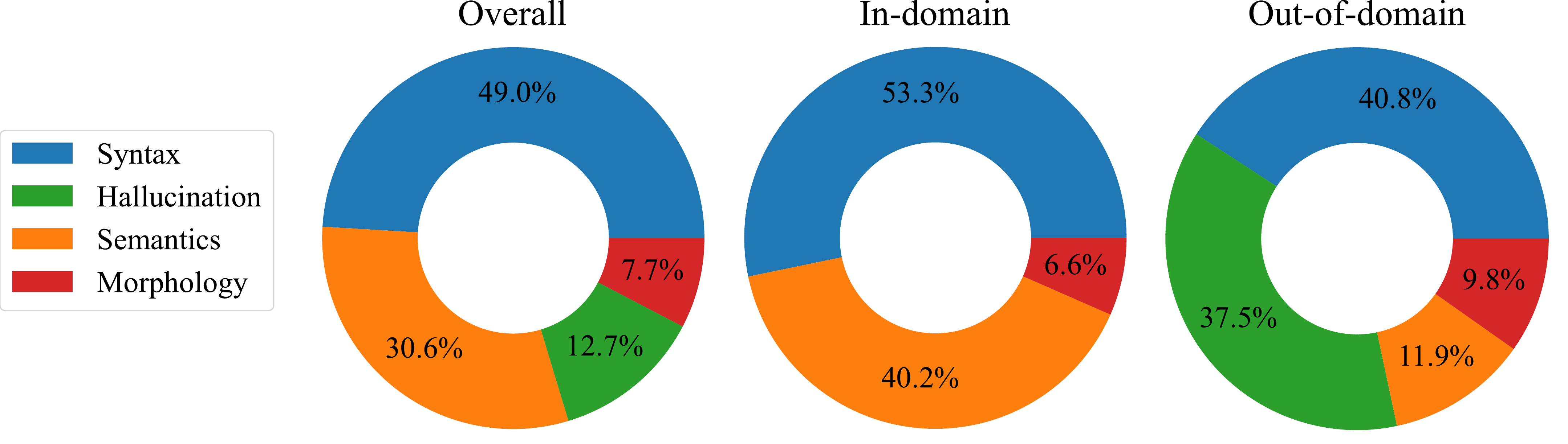}
  \caption{Distribution of violation categories in RuCoLA's unacceptable sentences.}
  \label{tax_dist}
  \vspace{-2ex}
\end{figure*}

\subsection{General Statistics}
\label{general_statistics}
\paragraph{Length and Frequency} The sentences in RuCoLA are filtered by the 4--30 token range with \texttt{razdel}\footnote{\href{http://github.com/natasha/razdel}{\texttt{github.com/natasha/razdel}}}, a rule-based Russian tokenizer. There are $11$ tokens in each sentence on average. We estimate the number of high-frequency tokens in each sentence according to the Russian National Corpus (RNC)\footnote{\href{http://ruscorpora.ru/new/en/}{\texttt{ruscorpora.ru/new/en}}} to control the word frequency distribution. It is computed as the number of frequently used tokens (i.e., the number of instances per million in RNC is higher than $1$) divided by the number of tokens in a sentence. We use a moderate frequency threshold $t \geqslant 0.6$ to keep sentences containing rare token units typical for some violations: non-existent or misderived words, incomplete translations, and others. The sentences contain on average $92$\% of high-frequency tokens.

\paragraph{Category Distribution} \autoref{tax_dist} shows the distribution of violation categories in RuCoLA. Syntactic violations are the most common in RuCoLA ($53.3$\% and $40.8$\% in the in-domain and out-of-domain sets). The in-domain set includes $40.2$\% of semantic and $6.6$\% of morphological violations, while the out-of-domain set accounts for $11.9$\% and $9.8$\%, respectively. Model hallucinations make up a percentage of $12.7$\% of the total number of unacceptable sentences.

\paragraph{Splits} The in-domain set of RuCoLA is split into train, validation and private test splits in the standard $80$/$10$/$10$ ratio ($7.9$k/$1$k/$1$k examples). The out-of-domain set is divided into validation and private test splits in a $50$/$50$ ratio ($1.8$k/$1.8$k examples). Each split is balanced by the number of examples per target class, the source type, and the violation category.

\begin{table*}[ht!]
\centering
\setlength{\tabcolsep}{6pt}
\begin{tabular}{@{}lcccccc@{}}
\toprule
\multirowcell{2}[-0.5ex][l]{\textbf{Baseline}} & \multicolumn{2}{c}{\textbf{Overall}} & \multicolumn{2}{c}{\textbf{In-domain}} & \multicolumn{2}{c}{\textbf{Out-of-domain}} \\ \cmidrule(lr){2-3} \cmidrule(lr){4-5} \cmidrule(lr){6-7}
 & \textbf{Acc.} & \textbf{MCC} & \textbf{Acc.} & \textbf{MCC} & \textbf{Acc.} & \textbf{MCC} \\ \midrule
\multicolumn{7}{c}{\textbf{Non-neural models}} \\ \midrule
Majority & 68.05\ {\small\pm \ 0.0} & 0.0\ {\small\pm \ 0.0} & 74.42\ {\small\pm \ 0.0} & 0.0\ {\small\pm \ 0.0} & 64.58\ {\small\pm \ 0.0} & 0.0\ {\small\pm \ 0.0} \\
Linear & 67.34\ {\small\pm \ 0.0} & 0.04\ {\small\pm \ 0.0} & 75.53\ {\small\pm \ 0.0} & 0.17\ {\small\pm \ 0.0} & 62.86\ {\small\pm \ 0.0} & -0.02\ {\small\pm \ 0.0} \\
\midrule \multicolumn{7}{c}{\textbf{Acceptability measures from LMs}} \\ \midrule
ruGPT-3 & 55.79\ {\small\pm \ 0.0} & 0.27\ {\small\pm \ 0.0} & 59.39\ {\small\pm \ 0.0} & 0.19\ {\small\pm \ 0.0} & 53.82\ {\small\pm \ 0.0} & 0.30\ {\small\pm \ 0.0}\\
\midrule \multicolumn{7}{c}{\textbf{Russian language models}} \\ \midrule
ruBERT & 75.9\ {\small\pm \ 0.42} & 0.42\ {\small\pm \ 0.01} & 78.82\ {\small\pm \ 0.57} & 0.4\ {\small\pm \ 0.01} & 74.3\ {\small\pm \ 0.71} & 0.42\ {\small\pm \ 0.01} \\
ruRoBERTa & \underline{80.8}\ {\small\pm \ 0.47} & \underline{0.54}\ {\small\pm \ 0.01} & \underline{83.48}\ {\small\pm \ 0.45} & \underline{0.53}\ {\small\pm \ 0.01} & \underline{79.34}\ {\small\pm \ 0.57} & \underline{0.53}\ {\small\pm \ 0.01} \\
ruT5 & 71.26\ {\small\pm \ 1.31} & 0.27\ {\small\pm \ 0.03} & 76.49\ {\small\pm \ 1.54} & 0.33\ {\small\pm \ 0.03} & 68.41\ {\small\pm \ 1.55} & 0.25\ {\small\pm \ 0.04} \\
\midrule
\multicolumn{7}{c}{\textbf{Cross-lingual models}} \\ \midrule
XLM-R & 65.73\ {\small\pm \ 2.33} & 0.17\ {\small\pm \ 0.04} & 74.17\ {\small\pm \ 1.75} & 0.22\ {\small\pm \ 0.03} & 61.13\ {\small\pm \ 2.9} & 0.13\ {\small\pm \ 0.05} \\
RemBERT & 76.21\ {\small\pm \ 0.33} & 0.44\ {\small\pm \ 0.01} & 78.32\ {\small\pm \ 0.75} & 0.4\ {\small\pm \ 0.02} & 75.06\ {\small\pm \ 0.55} & 0.44\ {\small\pm \ 0.01} \\ \midrule
Human & \textbf{84.08}  & \textbf{0.63} &\textbf{83.55} &\textbf{0.57} &\textbf{84.59} & \textbf{0.67} \\ \bottomrule
\end{tabular}
\caption{Results for acceptability classification on the RuCoLA test set. The best score is in bold, the second best one is underlined.}
\label{tab:Baseline-results}
\vspace{-2ex}
\end{table*}

\section{Experiments}

We evaluate several methods for acceptability classification ranging from simple non-neural approaches  to state-of-the-art cross-lingual models. %

\subsection{Performance Metrics} Following~\citet{warstadt-etal-2019-neural}, the performance is measured by the accuracy score (Acc.) and Matthews Correlation Coefficient (MCC, \citealp{Matthews1975ComparisonOT}). MCC on the validation set is used as the target metric for hyperparameter tuning and early stopping. We report the results averaged over ten restarts from different random seeds.

\subsection{Models}
\paragraph{Non-neural Models} We use two models from the \texttt{scikit-learn} library~\cite{pedregosa2011scikit} as simple non-neural baselines: a majority vote classifier, and a logistic regression classifier over tf-idf~\cite{Salton1973OnTS} features computed on word $n$-grams with the $n$-gram~range $\in[1;3]$, which results in a total of $2509$ features. For the linear model, we tune the $\ell_2$ regularization coefficient $C \in \{0.01, 0.1, 1.0\}$ based on the validation set performance.

\paragraph{Acceptability Measures}
Probabilistic measures allow evaluating the acceptability of a sentence while taking its length and lexical frequency into account~\cite{lau-etal-2020-furiously}.
There exist several different acceptability measures, such as PenLP, MeanLP, NormLP, and SLOR~\cite{lau-etal-2020-furiously}; we use PenLP due to its results in our preliminary experiments.
We obtain the PenLP measure for each sentence by computing its log-probability (computed as a sum of token log-probabilities) from the ruGPT3-medium\footnote{\href{https://hf.co/sberbank-ai/rugpt3medium_based_on_gpt2}{\texttt{hf.co/sberbank-ai/rugpt3medium}}} model.
PenLP normalizes the log-probability of a sentence $P(s)$ by the sentence length with a scaling factor~$\alpha$:
\begin{equation}
\textrm{PenLP}(s)=\frac{P(s)}{((5+|s|)(5+1))^\alpha}.
\end{equation}

After we compute the PenLP value of the sentence, we can predict its acceptability by comparing it with a specified threshold.
To find this threshold, we run 10-fold cross-validation on the train set: for each fold, we get the candidate thresholds on 90\% of the data by taking 100 points that evenly split the range between the minimum and maximum PenLP values. 
After that, we get the best threshold per fold by evaluating each threshold on the remaining 10\% of the training data.
Finally, we obtain the best threshold across folds by computing the MCC metric for each of them on the validation set.
\autoref{fig:penlp} in~\autoref{section:app_acceptability_measure} shows the distribution of scores for acceptable and unacceptable sentences, as well as the best PenLP threshold found in our experiments.

\paragraph{Finetuned Transformer Models} We use a broad range of monolingual and cross-lingual Transformer-based language models as our baselines. The monolingual LMs are ruBERT-base\footnote{\href{https://hf.co/sberbank-ai/ruBert-base}{\texttt{hf.co/sberbank-ai/ruBert-base}}} (178M trainable parameters), ruRoBERTa-large\footnote{\href{https://hf.co/sberbank-ai/ruRoberta-large}{\texttt{hf.co/sberbank-ai/ruRoberta-Large}}} (355M weights, available only in the large version), and ruT5-base\footnote{\href{https://huggingface.co/sberbank-ai/ruT5-base}{\texttt{hf.co/sberbank-ai/ruT5-base}}} (222M parameters). The cross-lingual models are XLM-R-base (\citealp{conneau-etal-2020-unsupervised}); 278M parameters) and RemBERT (\citealp{chung2020rethinking}; 575M parameters). All model implementations, as well as the base code for finetuning and evaluation, are taken from the Transformers library~\cite{transformers}. Running all experiments took approximately 126 hours on a single A100 80GB GPU.

All models except ruT5 are finetuned for 5 epochs with early stopping based on the validation set performance on each epoch. We optimize the hyperparameters of these models by running the grid search over the batch sizes $\{32, 64\}$, the learning rates $\{10^{-5}, 3\cdot 10^{-5}, 5\cdot 10^{-5}\}$ and the weight decay values $\{10^{-4}, 10^{-2}, 0.1\}$. We fine-tune ruT5 for 20 epochs (also using early stopping) with the batch size of 128; the search space is $\{10^{-4}, 10^{-3}\}$ for the learning rate and $\{0, 10^{-4}\}$ for the weight decay respectively. 

The classification task for ruT5 is framed as a sequence-to-sequence problem: we encode the ``acceptable'' label as ``yes'' and the ``unacceptable'' one as ``no''. The model takes the sentence as its input and generates the corresponding label. We interpret all strings that are not equal to ``yes'' or ``no'' as predictions of the ``unacceptable'' class.

\subsection{Human Evaluation} \label{subsection:human}

We conduct a human evaluation on the entire in-domain test set and 50\% of the out-of-domain test set. The pay rate is on average \$$6.3$/hr, and the task design is similar to \textbf{Stage 1} in \S\ref{paragraph:stage_one} (see also \autoref{tab:binary},  Appendix~\ref{appendix:instructions}) with a few exceptions. In particular, (i) we remove the ``Not confident'' answer option,  (ii) the annotators are 16 undergraduate BA and MA students in philology and linguistics from Russian universities, and (iii) the votes are aggregated using the method by~\citet{dawid_skene}, which is available directly from the Toloka interface. The average quality rate on the control tasks exceeds $75$\%.

\section{Results and Analysis}
\vspace{-2pt}

\label{section:results}
\autoref{tab:Baseline-results} outlines the acceptability classification results. Overall, we find that the best-performing ruRoBERTa model still falls short compared to humans and that different model classes have different cross-domain generalization abilities. Below, we discuss our findings in detail.

\begin{table*}[!hbp]
\vspace{-2ex}
\centering
\small
\setlength{\tabcolsep}{8pt}
\begin{tabular}{lccccc}
\toprule
\textbf{Method} &
\textbf{Acceptable} &
\textbf{Hallucination} &
\textbf{Morphology} & \textbf{Semantics} & \textbf{Syntax} \\
 \midrule
 \multicolumn{6}{c}{\textbf{Non-neural models}} \\ \midrule
Majority & 100.0\ {\tiny\pm \ 0.0} & 0.0\ {\tiny\pm \ 0.0} & 0.0\ {\tiny\pm \ 0.0} & 0.0\ {\tiny\pm \ 0.0} & 0.0\ {\tiny\pm \ 0.0} \\
Linear & 96.5\ {\tiny\pm \ 0.0} & 3.3\ {\tiny\pm \ 0.0} & 3.8\ {\tiny\pm \ 0.0} & 3.4\ {\tiny\pm \ 0.0} & 7.4\ {\tiny\pm \ 0.0} \\
\midrule \multicolumn{6}{c}{\textbf{Acceptability measures from LMs}} \\ \midrule
ruGPT-3 & 36.5\ {\tiny\pm \ 0.0} & 77.4\ {\tiny\pm \ 0.0} & 68.8\ {\tiny\pm \ 0.0} & 63.1\ {\tiny\pm \ 0.0} & 77.6\ {\tiny\pm \ 0.0} \\
\midrule \multicolumn{6}{c}{\textbf{Russian language models}} \\ \midrule
ruBERT & 87.7\ {\tiny\pm \ 1.9} & 62.6\ {\tiny\pm \ 5.0} & 30.8\ {\tiny\pm \ 3.5} & 33.8\ {\tiny\pm \ 2.8} & 55.2\ {\tiny\pm \ 3.6} \\
ruRoBERTa & 91.5\ {\tiny\pm \ 1.2} & 63.4\ {\tiny\pm \ 4.5} & 44.4\ {\tiny\pm \ 4.0} & 37.1\ {\tiny\pm \ 3.1} & 66.8\ {\tiny\pm \ 2.9} \\
ruT5 & 89.9\ {\tiny\pm \ 3.6} & 35.4\ {\tiny\pm \ 6.5} & 17.0\ {\tiny\pm \ 4.1} & 20.6\ {\tiny\pm \ 6.2} & 37.0\ {\tiny\pm \ 4.6} \\
\midrule \multicolumn{6}{c}{\textbf{Cross-lingual models}} \\ \midrule
XLM-R & 79.9\ {\tiny\pm \ 6.2} & 39.7\ {\tiny\pm \ 9.6} & 29.4\ {\tiny\pm \ 11.4} & 17.0\ {\tiny\pm \ 4.6} & 42.9\ {\tiny\pm \ 7.1} \\
RemBERT & 85.6\ {\tiny\pm \ 1.7} & 64.6\ {\tiny\pm \ 4.1} & 37.8\ {\tiny\pm \ 3.2} & 35.3\ {\tiny\pm \ 4.2} & 64.4\ {\tiny\pm \ 2.8} \\ \midrule
Human & 87.7\ {\tiny\pm \ 0.0} & 84.5\ {\tiny\pm \ 0.0} & 81.5\ {\tiny\pm \ 0.0} & 57.1\ {\tiny\pm \ 0.0} & 80.1\ {\tiny\pm \ 0.0} \\
\bottomrule
\end{tabular}
\vspace{-1ex}
\caption{Per-category recall on the RuCoLA test set.}
\label{tab:error-analysis}
\end{table*}

\vspace{-2pt}
\subsection{Acceptability Classification}

ruRoBERTa achieves the best overall performance among the trained methods, which is nine points behind the human baseline in terms of overall MCC score. The second-best model is RemBERT, followed by ruBERT, with scores of 10\% and 12\% below ruRoBERTa, respectively. ruT5 and ruGPT-3 + PenLP perform similarly in terms of MCC, although the accuracy of ruT5 is significantly higher. XLM-R achieves the worst performance among finetuned neural models, and the majority vote and logistic regression classifiers have near-zero MCC scores.

We observe that the best models perform similarly on the in-domain and out-of-domain sets with an absolute difference of 0 to 0.04 in terms of MCC. However, the performance gap for other LMs is more prominent. RuT5, XLM-R, and the logistic regression drop by approximately 10 points, whereas the ruGPT-3 + PenLP performance increases. 
RuT5 and XLM-R have fewer parameters than RuRoBERTa and RemBERT, and smaller models tend to rely more on surface-level cues that poorly transfer to a different domain~\cite{niven-kao-2019-probing}.   
The increase in quality for out-of-domain set for PenLP is due to ruGPT-3 assigning consistently lower probabilities to generated sentences. Thus, the PenLP values are skewed to the left for unacceptable sentences (see~\autoref{fig:penlp} in~\autoref{section:app_acceptability_measure}).

The human performance is higher on the out-of-domain dataset, which can be attributed to the ``unnaturalness'' of machine-specific features, e.g., hallucinations, nonsense, and repetitions~\cite{nucleus,typical}. The presence of such generated text properties may directly indicate the unacceptability of a sentence.

Finally, we observe that the monolingual models tend to outperform or perform on par with the cross-lingual ones. We attribute this to the size of pre-training data in Russian, which can be five times larger (for ruRoBERTa compared to XLM-R). The size and the quality of the pre-training corpora may directly affect learning the language properties.
It might be possible to test this hypothesis by comparing a series of monolingual and cross-lingual LMs pre-trained on the datasets of varying sizes and studying how scaling the data impacts the acquisition of grammatical phenomena~\cite{zhang-etal-2021-need}; however, such a study is beyond the scope of our work.

\vspace{-6pt}
\subsection{Error Analysis}
\vspace{-2pt}
\label{sect:error_analysis}
To understand the similarities and differences between classification patterns of models and humans, we conduct an error analysis of all evaluated methods on the in-domain test set. Specifically, we study the proportion of incorrectly classified examples in each group (acceptable sentences and three violation categories).

    The main quantitative results of our analysis are shown in~\autoref{tab:error-analysis}. Our manual study of $250$ examples misclassified by all methods reveals that sentences with non-specific indefinite pronouns, adverbials, existential constructions, and phrases with possessive prepositions are the most challenging for models and human annotators. We also find that monolingual and cross-lingual LMs tend to judge sentences with the ungrammatical agreement and government as acceptable (e.g.,\textit{*Kakim \textcolor{cb-burgundy}{vami} viditsja buduschee strany?} ``How do you see \textcolor{cb-burgundy}{your} see the future of the country?''). Humans make mistakes in long sentences with comparative and subordinate clauses and prepositional government. Another observation is that LMs are not sensitive to morphological violations, such as misderived comparative forms (\textit{*Oni v'ehali v bor, i zvuk kopyt stal \textcolor{cb-burgundy}{zvonchee}.} ``They drove into the forest, and the sound of hooves became \textcolor{cb-burgundy}{louderer}.''), ungrammatical word-building patterns, and declension of numerals. Finally, most acceptability classifiers achieve high recall on hallucinated sentences, which confirms a practical application potential for classifiers trained on RuCoLA.

\vspace{-6pt}
\subsection{Effect of Length}
\vspace{-2pt}

\begin{figure}[ht!]
\includegraphics[width=\linewidth]{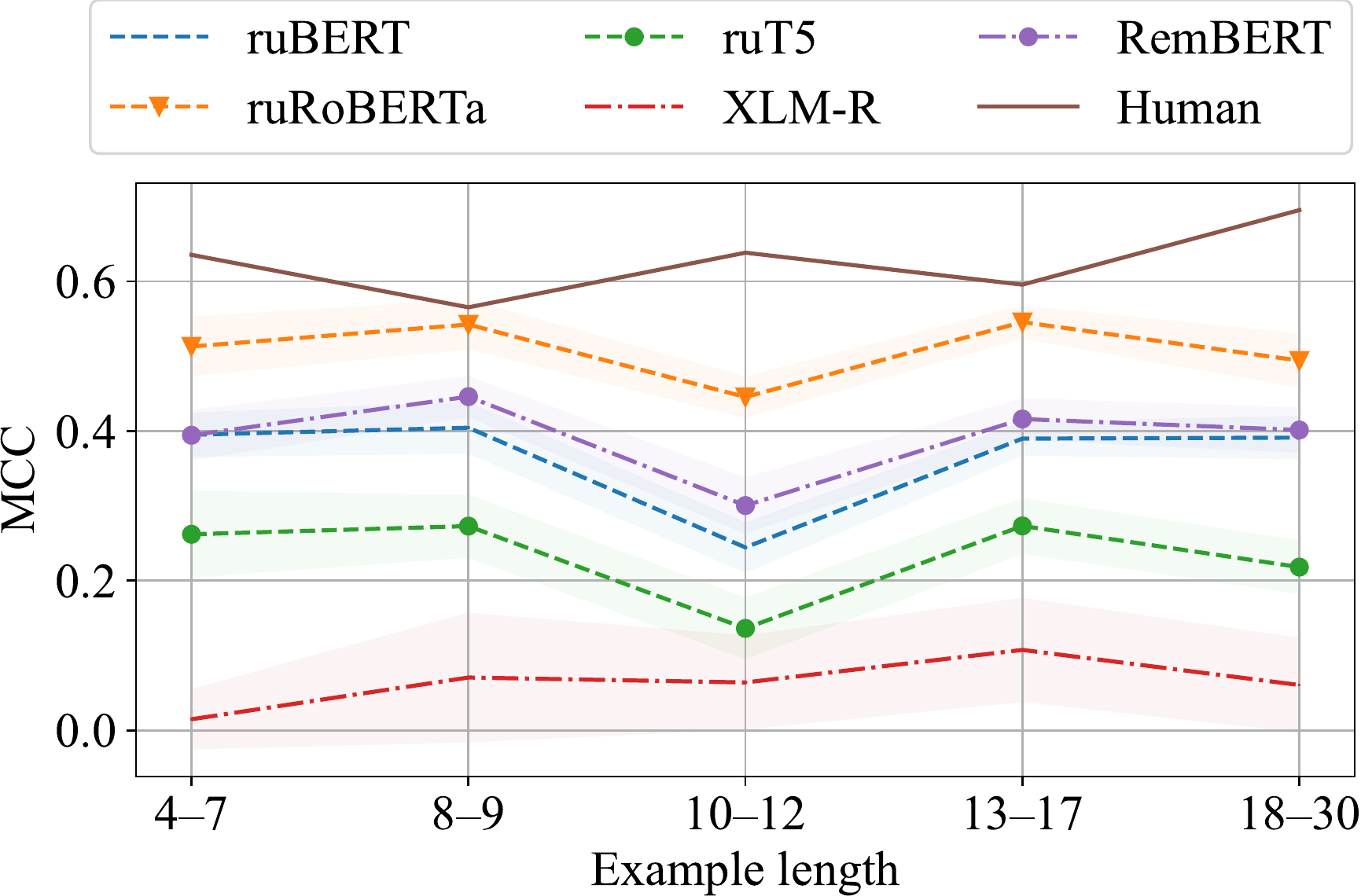}
\vspace{-3ex}
\caption{Results on the RuCoLA test set grouped by five quintiles of the sentence length.}
\label{fig:results_by_length}
\vspace{-2ex}
\end{figure}

We analyze the effect of sentence length on the acceptability classification performance by dividing the test set into five length groups of equal size. The results are displayed in \autoref{fig:results_by_length}. The general trend is that the behavior of performance is consistent across all methods. However, while the model performance is unstable and slightly degrades as the length increases, the human annotators outperform the language models on all example groups. Overall, our results are consistent with the findings of~\citet{warstadt2019linguistic}.

To discover the reason behind the increase in quality of automatic methods for sentences of $13$--$17$ tokens, we manually studied a subset of $50$ sentences misclassified by ruRoBERTa for each bucket, which amounts to $250$ examples in total. We observed that the domain distribution and the error type distribution vary between the length quintile groups, which could explain the differences in model performance for these groups. Specifically, the third group ($10$--$12$ tokens) contains sentences with ungrammatical agreement and government or violated argument structure, which are difficult for the models (see Section~\ref{sect:error_analysis}). In turn, the fourth quintile interval ($13$--$17$ tokens) has more out-of-domain examples of hallucinations, which are easier to detect both for humans and ML-based methods.

\section{Cross-lingual Transfer}
\vspace{-1ex}

\label{sect:crosslingual}
Given the availability of acceptability classification corpora in other languages, one might be curious about the possibility of knowledge transfer between languages for this task. This is particularly important in the case of estimating sentence acceptability for low-resource languages, which are an important focus area of NLP research~\cite{hedderich-etal-2021-survey}. However, the nature of the task makes successful transfer an open question: for instance, specific grammar violations in one language might not exist in another.

With this in mind, we explore the zero-shot cross-lingual transfer scenario, in which the training and validation datasets are provided in one language and the test data in a different one. We use four multilingual models: mBERT~\cite{devlin-etal-2019-bert}, XLM-R\textsubscript{Base}, XML-R, and RemBERT.  We study the transfer between three datasets: CoLA, ItaCoLA, and RuCoLA, containing examples in English, Italian and Russian, respectively. As shown in Table~\ref{tab:comparison}, all datasets have similar sizes.

Due to the space constraints, we defer a detailed description of the experimental setup and results to \autoref{section:appendix_crosslingual}; here, we overview the key findings of this study. Specifically, we find that the monolingual scenarios outperform cross-lingual transfer by a large margin, which confirms and extends the results of~\citet{trotta-etal-2021-monolingual-cross}. 
Also, we observe that RemBERT performs best in monolingual and cross-lingual setups.
For the in-domain set, we observe a cross-lingual transfer gap: there is little difference in language to transfer from, and for RuCoLA, the zero-shot results are as poor as those of a linear ``monolingual'' classifier.
However, the cross-lingual setup performs on par with the monolingual setup for out-of-domain data.

\section{Conclusion and Future Work}

This work introduces RuCoLA, the first large-scale acceptability classification corpus in the Russian language. The corpus consists of more than 13.4k sentences with binary acceptability judgments and provides a coarse-grained annotation of four violation categories for 3.7k unacceptable sentences. RuCoLA covers two types of data sources: linguistic literature and sentences produced by generative models. Our design encourages NLP practitioners to explore a wide range of potential applications, such as benchmarking, diagnostic interpretation of LMs, and evaluation of language generation models.
We conduct extensive experimental evaluation by training baselines that cover a broad range of models. Our results show that LMs fall behind humans by a large margin. Finally, we explore the cross-lingual generalization capabilities of four cross-lingual Transformer LMs across three languages for acceptability classification. The preliminary results show that zero-shot transfer for in-domain examples is hardly possible, but the discrepancy between monolingual and cross-lingual training results for out-of-domain sentences is less evident.
 
In our future work, we plan to explore the benefits and limitations of RuCoLA in the context of applying acceptability classifiers to natural language generation tasks. Another direction is to augment the in-domain and out-of-domain validation sets with fine-grained linguistic annotation for nuanced and systematic model evaluation. In the long run, we hope to provide valuable insights into the process of grammar acquisition by language models and help foster the application scope of linguistic acceptability.
 
\section{Limitations}
\paragraph{Data Collection} Acceptability judgments datasets require a source of unacceptable sentences. Collecting judgments from linguistic literature has become a standard practice replicated in multiple languages. However, this approach has several limitations. First, many studies raise concerns about the reliability and reproducibility of acceptability judgments~\citep[e.g.,][]{gibson2013need,culicover2010quantitative,sprouse2013empirical,linzen2018reliability}. Second, the linguists' judgments may limit data representativeness, as they may not reflect the errors that speakers tend to produce~\cite{dkabrowska2010naive}. Third, enriching acceptability judgments datasets is time-consuming, while creating new ones can be challenging due to limited resources, e.g., in low-resource languages.

\paragraph{Expert vs. Non-expert} One of the open methodological questions on acceptability judgments is whether they should be collected from expert or non-expert speakers. On the one hand, prior linguistic knowledge can introduce bias in reporting judgments~\cite{gibson2010weak}. On the other hand, expertise may increase the quality of the linguists' judgments over the ones of non-linguists~\citep[see a discussion by][]{schutze2013judgment}. At the same time, the latter tend to be influenced by an individual's exposure to ungrammatical language use~\cite{dkabrowska2010naive}. Recall that our in-domain examples and their acceptability labels are manually drawn from linguistic literature, while the out-of-domain set undergoes two stages (\S\ref{annotation_machine}):
\begin{enumerate}[noitemsep,leftmargin=2.em]
    \item \textbf{Stage 1: Acceptability Judgments} --- collecting acceptability labels from non-expert speakers;
    \item \textbf{Stage 2: Violation Categories} --- validation of the acceptability labels from \textbf{Stage 1} and fine-grained example annotation by their violation category by expert speakers.
\end{enumerate}

The objective of involving students with a linguistic background is to maximize the annotation quality. We follow~\citet{warstadt-etal-2019-neural} and report the students' evaluation results as the human baseline in this paper. Human evaluation through crowdsourcing~\cite{nangia-bowman-2019-human} is left for future work.

\paragraph{Fine-grained Annotation} The coarse-grained annotation scheme of the RuCoLA's unacceptable sentences relies on four major categories. While the annotation can be helpful for model error analysis, it limits the scope of LMs' diagnostic evaluation concerning linguistic and machine-specific phenomena~\cite{warstadt2019linguistic}.

\paragraph{Distribution Shifts} Many studies have discussed the role of lexical frequency in acceptability judgments~\cite{myers2017acceptability}. In particular, LMs can treat frequent patterns from their pre-training corpora as acceptable and perform poorly on rare or unattested sentences with low probabilities~\cite{marvin-linzen-2018-targeted,park2021deep,linzen2021syntactic}. Although we aim to control the number of high-frequency tokens in the RuCoLA's sentences (\S\ref{general_statistics}), we assume that potential word frequency distribution shift between LMs' pre-training corpora and our corpus can introduce bias in the evaluation. Furthermore, linguistic publications represent a specific domain as the primary source of acceptability judgments. On the one hand, it can lead to a domain shift when using RuCoLA for practical purposes. On the other hand, we observe moderate acceptability classification performance on the out-of-domain test, which spans multiple domains, ranging from subtitles to Wikipedia.

\section{Ethical Considerations} Responses of human annotators are collected and stored anonymously. The average annotation pay rate exceeds the hourly minimum wage in Russia twice or four times, depending on the annotation project. The annotators are warned about potentially sensitive topics in data (e.g., politics, culture, and religion).

RuCoLA may serve as training data for acceptability classifiers, which may benefit the quality of generated texts~\cite{batra-etal-2021-building}. We recognize that such improvements in text generation may lead to misuse of LMs for malicious purposes~\cite{weidinger2021ethical}. However, our corpus can be used to train adversarial defense and artificial text detection models. This paper introduces a novel dataset for \textbf{research and development needs}, and the potential negative uses are not lost on us.

\section{Acknowledgements}
Alena Pestova was supported by the framework of the HSE University Basic Research Program. Max Ryabinin was supported by the grant for research centers in the field of AI provided by the Analytical Center for the Government of the Russian Federation (ACRF) in accordance with the agreement on the provision of subsidies (identifier of the agreement 000000D730321P5Q0002) and the agreement with HSE University No. 70-2021-00139. The data annotation effort was supported by the Toloka Research Grants program.

\bibliography{custom,anthology}
\bibliographystyle{acl_natbib}

\clearpage
\newpage

\appendix

\section{Examples}
\label{appendix:examples}
This appendix provides approximately 50 examples of the RuCoLA's sentences appearing in the in-domain and out-of-domain sets and corresponding fine-grained phenomena.

\subsection{In-domain Set}
\paragraph{Morphology}
\ex. \textbf{Word derivation}
\a. Comparatives
    \a. *Litso ― dlin'she, chem nado, a telo ― koroche. (``The face is longer than it should be, and the body is shorter.'')
    \b. *A sapogi ja vam blestjashhie prinesu, eshhjo krasivshe vashih! (``And I'll bring you shiny boots, even more beautiful than yours!'')
    \z.
\b. Word-building patterns
    \a. *Ljudej rugajut ili hvaljat za ihnie dela, a ne za nacional'nost'. (``People are scolded or praised for their deeds, not for their nationality.'')
    \b. *Vdobavok, v koridore bylo holodno, i mal'chik sovsem ozjabnul. (``In addition, it was cold in the corridor, and the boy was completely chilled.'')
    \b. *Zarubezhnym kollegam predlozhili propoverjat' rezul'taty. (``Foreign colleagues were invited to check the results.'')
    \z.
\b. Declension of numerals
    \a. *Delo sostoit v tom, chto ``Nezhnyj vestnik'' rashoditsja v vos'm'justah kopijah ezhenedel'no, po tridcati kopeek.
    \b. *My darim podarok kazhdyj pjat'sotyj zakaz, opredeljaemyj po nomeru nakladnoj. (``We give a gift every five hundredth order, determined by the invoice number.'')
    \z.

\paragraph{Syntax}
\ex. \textbf{Copular constructions}
    \a. Vse utro on byl razdrazhitelen. (``He had been irritable all morning.'')
    \b. *U nas sejchas est' dozhd'. (``It is raining now.'')
    \z.

\ex. \textbf{Word order}
\a. Subordinate clauses
    \a. *Den' goroda, kotorom ja zhivu v. (``The day of the city I live in.'')
    \b. Ja prines dokumenty, chtoby mne ne byt' na sude goloslovnym. (``I brought the documents so that I wouldn't be unfounded at the trial.'')
    \z.
\b. Coordinate clauses and constructions
    \a. *I on podnjal trubku, ja pozvonil Vane. (``And he picked up the phone, I called Vanya.'')
    \b. *Ona to stihi chitaet, kartiny to pokazyvaet. (``She either reads poetry, shows or pictures.'')
    \z.

\ex. \textbf{Agreement}
\a. Number agreement
    \a. *Devochki, davaj zajdem v magazin! (``Girls, let's go to the store!'')
    \b. *Te, kto nazyvajut sebja patriotami, dolzhen horosho znat' rodnoj jazyk. (``Those who call themselves patriots should know their native language well.'')
    \z.
\b. Case agreement
    \a. *Ego ot'ezd za granicu vsem vosprinimalsja kak pobeg. (``His departure abroad was perceived by everyone as an escape.'')
    \b. *Na melkovodnyh uchastkah rastitel'nost' obrazuet peremychki, razdeljajushhimi ozero na otdel'nye pljosy. (``In shallow areas, vegetation forms bridges dividing the lake into separate stretches.'')
    \b. *Nikogo ne bylo kholodno. (``No one was cold.'')
    \z.

\ex. \textbf{Verb transitivity}
\a. Intransitive verbs with prepositional phrases
    \a. *Na kazhdoj dorozhke bezhalo po sportsmenu (``There was an athlete running on each track.'')
    \b. *Po rebenku sdelali sebe buterbrody. (``For the child made themselves sandwiches.'')
    \z. 
\b. Transitive verbs with impersonal clauses or sentential actants
    \a. U Nonny gorelo lico, ee dazhe znobilo ot volnenija. (``Nonna's face was burning, she was even shivering with excitement.'')
    \b. *On znaet, chto polk perebrosjat na drugoj uchastok fronta i drugie plany komandovanija. (``He knows that the regiment will be transferred to another sector of the front and other plans of the command.'')
    \z.

\ex. \textbf{Coordination and subordination}
\a. Constructions with subordinate and infinitive clauses
    \a. *Nam uzhe bylo izvestno, chto on priekhal i drugie fakty.
    (``We already knew that he had arrived and other facts.'')
    \b. *I chto on byl razbit, bylo zamecheno vsemi. (``And that it was broken, everything was noticed.'')
    \z.
\b. Coordinate clauses with dative constructions
    \a. *Mne vystupat' sledujushhim i uzhe napomnili ob jetom. (``I will be the next to speak and have already been reminded of this.'')
    \b. Mne soobshhili ob jetih planah, i oni ponravilis'. (``I was informed about these plans, and I liked them.'')
    \z.

\paragraph{Semantics}
\ex. \textbf{Non-specific indefinite pronouns}
    \a. *Khorosho, chto on kupil chto-nibud'. (``It's good that he bought something.'')
    \b. *Kakogo-nibud' reshenija on ne prinjal. (``He didn't make any decision.'')
    \b. *Ja ne ljublju kogo-libo. (``I don't love anyone.'')
    \z.

\ex. \textbf{Tense}
    \a. *Zavtra my slyshim operu. (``Tomorrow we hear the opera'')
    \z.

\ex. \textbf{Aspect}
    \a. *Zavtra budem ezdit' vo Vneshtorgbank. (``Tomorrow we will go to Vneshtorgbank.'')
    \z.

\ex. \textbf{Negation or negative concord}
    \a. *Nikto ego videl? (``Has no one seen him?'')
    \b. *On ne byl tam i razu. (``He hasn't been there once.'')
    \z.
    
\ex. \textbf{Existential constructions}
    \a. *Sushhestvujut zajavlenija ot postradavshikh.(``There are statements from victims.'')
    \z.

\setcounter{ExNo}{0}

\subsection{Out-of-domain Set}
\paragraph{Morphology}

\ex. \textbf{Nonce words}
    \a. *I ja sygral pervoe dvizhenie betovennogo violetovogo koncerta. (``And I played the first movement of the beethoven violette concerto.'')
    \b. *Rastenie harakterno dlja stepi i sil'vostepi na ravninah i na plato Moldavii na severe. (``The plant is characteristic of the steppe and the silvosteppe on the plains and on the plateau of Moldova in the north.'')
    \b. *Aviakompanijam razreshili ispol'zovat' servis ``onechuckle'' dlja zakaza samyh populjarnyh aviamar. (``Airlines were allowed to use the "onechuckle" service to order the most popular aviamars.'')
    \b. *Ona risuet horosho i mechtaet stat' hudozhn'ej. (``She draws well and dreams of becoming an artistrone.'')
    \b. *Dlja nihsudarstvennyh organizacij razrabotali metod analiza bjudzhetov dlja razreshenija konfliktov s zhenshhinami.  (``A method of budget analysis for resolving conflicts with women has been developed for themgovernmental organizations.'')
    \z.

\paragraph{Syntax}

\ex. \textbf{Agreement}
\a. Person
    \a.  *On ostalsja nevredimoj i v moroze. (``He remained unharmed and in the cold.'')
    \z. 
\b. Case
    \a. *Vospol'zujtes' prokat avtomobilej i 24-chasovoj priem, chtoby vy mogli ispytat' svoj prebyvanie, kak vy hotite. (``Take advantage the car rental and 24-hour reception so you can experience your stay the way you want.'')
    \z.

\ex. \textbf{Subordination and coordination}
    \a. *Jeto to, chto oni prishli k ponimaniju, chto samoe vazhnoe, chemu deti dolzhny nauchit'sja, jeto harakter. (``This is what they have come to understand that the most important thing children need to learn this is character.'')
    \z.

\ex. \textbf{Ellipsis}
    \a. *Bolee 30 uchenyh zashhitili kandidatskie dissertacii pod rukovodstvom. (``More than 30 scientists defended their PhD theses under the supervision of.'')
    \z.

\paragraph{Hallucinations}

\ex. \textbf{Nonsensical sentences}
    \a. *Soobshhenie s nej tol'ko po peshke. (``Messaging her is only possible by a pawn.'')
    \b. *Futbolist ``Liverpulja'' vpervye podpisal pervyj god porazhenija kolena. (``The Liverpool footballer has signed for the first time in the first year of defeating his knee.'')
    \b. *I vse po vsej biblioteke raznye predmety, raznye prostranstva. (``And all throughout the library are different objects, different spaces.'')
    \z.

\ex. \textbf{Irrelevant repetitions}
    \a. *Dlja jetoj programmy byli provedeny dva programmy.  (``For this program two programs were conduncted.'')
    \b. *Posylki pojavlenija product placement v kinematografe u brat'ev Ljum'er pojavilis' uzhe u brat'eva Ljum'era. (``The premises of the appearance of the product placement in the cinema of the Lumiere brothers have already appeared in the Lumiere brothers.'')
    \z.

\paragraph{Semantics}
\ex. \textbf{Semantics}
    \a. *Poberezh'ja Ivanova i Kohma proshli na severe. (``The coasts of Ivanovo and Kokhma passed in the north.'')
    \b. *Prezident zajavil, chto u Rossii dostatochno sil dlja provedenija profilakticheskih zabastovok. (``The President said that Russia has enough forces to carry out preventive strikes.'')
    \b. *Torgovlja real'nymi den'gami na virtual'nom rynke vyrosla, chtoby stat' mnogomillionnoj industriej dollarov.  (``Real money trading in the virtual market has grown to become a multi-million dollar industry.'')
    \b. *On vnov' zavershil nokautom pretendenta v vos'mom raunde.  (``He again finished by knocking out the challenger in the eighth round.'')
    \z.

\begin{table*}

\section{Annotation Protocols} \label{appendix:annotation_protocol}
\subsection{Instructions}
\label{appendix:instructions}
\scriptsize

\begin{minipage}[t]{.43\linewidth}
\par\noindent\rule{\textwidth}{1pt}

\vspace{.5cm}

\textbf{Task}
\vspace{0.05cm}
\begin{itemize}[noitemsep,topsep=0.1pt]
    \item Your task is to define whether a given sentence is appropriate or contains any violations (one would not say or write like this).
    \item Choose ``Yes'' if the sentence contains one or more violations.
    \item Choose ``No'' if the sentence is appropriate (you would say like this).
    \item Choose ``Not confident'' if you have doubts.
    \item If there are any typos, please state them in the box.
\end{itemize}

\vspace{0.2cm}
\textbf{Examples of violations}
\vspace{0.05cm}

\begin{itemize}[noitemsep,topsep=0.1pt]
    \item Number disagreement: \textit{``Podrobnosti dela \textcolor{cb-burgundy}{neizvestno}.''} (``No details of this case \textcolor{cb-burgundy}{is} available.'')

    \item Semantic collisions: \textit{``\textcolor{cb-burgundy}{V etom godu} zhenschiny vyshly zamuzh vo vtoroy raz \textcolor{cb-burgundy}{26 iyunya 1989 goda}.''
    } (``\textcolor{cb-burgundy}{This year} the women got married the second time  \textcolor{cb-burgundy}{on the 26th of June in 1989}.'')
    
    \item Nonsensical repetitions: \textit{``Eto \textcolor{cb-burgundy}{moya sem`ya moya sem'ya}.''} (``It is \textcolor{cb-burgundy}{my family my family}.'')
\end{itemize}

\vspace{0.2cm}
\textbf{Annotation examples}
\vspace{0.05cm}

\begin{itemize}[noitemsep,topsep=0.1pt]
    \item ``Yes`` (the given sentence contains one or more violations): \textit{``Ya dolzhen poiti \textcolor{cb-burgundy}{s velichiem}, chtoby prostit' eyoh.}
    (``I should go \textcolor{cb-burgundy}{with greatness} to forgive her.'')
    
    \item ``No'' (the given sentence is acceptable): \textit{``Skol'ko chasov v den' vy rabotaete?''} (``How many hours a day do you work?'')
\end{itemize}

Please check the task before submission. \\
Thank you!

\par\noindent\rule{\textwidth}{0.8pt}

\vspace{0.05cm}
\textbf{Example of web interface}
\vspace{0.1cm}

Does the sentence contain \textbf{violations}?
\vspace{0.1cm}

\colorbox{Gray}{This is a toy example.}

\begin{itemize}[noitemsep,topsep=0pt]
    \item[\radiobutton] Yes 
    \item[\radiobutton] No
    \item[\radiobutton] Not confident

\end{itemize}

If there are any typos, please state them below:

\fbox{
    \begin{minipage}{0.4\textwidth}
    \parbox{0.4\textwidth}{
        \centering
        \tiny
     }
     \end{minipage}
}

\vspace{0.1cm}
Please check the task once again. Thank you!

\par\noindent\rule{\textwidth}{1pt}

\caption{A shortened version of the instruction given to crowd-sourced annotators for judging the acceptability of machine-generated sentences (\textbf{Stage 1: Acceptability Judgments}; \S\ref{paragraph:stage_one}). The instruction is translated for illustration purposes.}

\label{tab:binary}
\end{minipage}%
\hspace{0.1\textwidth}%
\begin{minipage}[t]{.43\linewidth}

\par\noindent\rule{\textwidth}{1pt}

\vspace{.5cm}

\textbf{Task}
\begin{itemize}[noitemsep,topsep=0.1pt]
    \item Your task is to select all appropriate violation categories under which a given sentence falls: Morphology, Syntax, Semantics, Hallucinations, or Other.

    \item Choose ``No violations'' if the sentence is acceptable.
    \item Choose ``Not confident'' if you have doubts.
    \item If there are any typos, please state them in the box.
    \item If any questions or doubts, contact us in the chat.  
\end{itemize}

\vspace{0.2cm}
\textbf{Examples}
\vspace{0.05cm}
\begin{itemize}[noitemsep,topsep=0.1pt]
    \item Morphology
    \begin{itemize}[noitemsep,topsep=0.1pt]
        \item Non-existent words: \textit{``Eto semiduymovyi \textcolor{cb-burgundy}{heturpin}.''} (``It is a seven-inch \textcolor{cb-burgundy}{heturpin}.'')
        \item Misderivation: \textit{``Ona vyglyadit \textcolor{cb-burgundy}{krasivshe}.''} (``She looks \textcolor{cb-burgundy}{more beatufiuler}''.)
    \end{itemize}

    \item Syntax
    \begin{itemize}[noitemsep,topsep=0.1pt]
        \item Agreement violation: \textit{``Oni schitali ego \textcolor{cb-burgundy}{talantlivymi}.''} (``They considered him to be \textcolor{cb-burgundy}{talented}.'')
        \item Word order: \textit{``Plan Mashe Sashi prodat' kvartiru.''} (``Plan Masha's Sasha to sell a flat.'')
    \end{itemize}
    
    \item Semantics
    \begin{itemize}[noitemsep,topsep=0.1pt]
        \item Semantic properties of the predicate: ``Ty kogda-nibud' \textcolor{cb-burgundy}{nakhodilsya} v Moskve?'' (``Have you ever \textcolor{cb-burgundy}{been} to Moscow?'')
    \end{itemize}
    
    \item Hallucinations
    \begin{itemize}[noitemsep,topsep=0.1pt]
        \item Incomplete translation or input copying: \textit{``Ya rad, shto \textcolor{cb-burgundy}{you heard} o \textcolor{cb-burgundy}{Margaret Thatcher}.''} (``I am glad \textcolor{cb-burgundy}{you heard} about \textcolor{cb-burgundy}{Margaret Thatcher}.``)
        
        \item Repetitive content: \textit{``Eto \textcolor{cb-burgundy}{moya sem`ya moya sem'ya}.''} (``It is \textcolor{cb-burgundy}{my family my family}.'')
        
    \end{itemize}
    
\end{itemize}

\par\noindent\rule{\textwidth}{0.8pt}
\vspace{0.05cm}
\textbf{Example of web interface}
\vspace{0.1cm}

Select all appropriate violation categories.
\vspace{0.1cm}

\colorbox{Gray}{This is a toy example.}

\begin{itemize}[noitemsep,topsep=0pt]
    \item[$\square$] Morphology 
    \item[$\square$] Syntax
    \item[$\square$] Semantics
    \item[$\square$] Hallucinations
    \item[$\square$] Other
    \item[$\square$] Not confident
    \item[$\square$] No violations
\end{itemize}

If there are any typos, please state them below:

\fbox{
    \begin{minipage}{0.4\textwidth}
    \parbox{0.4\textwidth}{
        \centering
        \tiny
     }
     \end{minipage}
}

\vspace{0.2cm}
Please check the task before submission. \\
Thank you!

\par\noindent\rule{\textwidth}{1pt}

\caption{A shortened version of the instruction given to students for validation and coarse-grained annotation of the unacceptable machine-generated sentences (\textbf{Stage 2: Violation Categories}; \S\ref{paragraph:stage_two}). The instruction is translated for illustration purposes.}
\label{table:multilabel}

\end{minipage}

\end{table*}

\clearpage
\newpage

\begin{table*}[th!]
\setlength{\tabcolsep}{4pt}
\small
\centering
\begin{tabular}{lcccccccc}
\toprule
 & \textbf{\# annotators} & \textbf{Pay rate}  & \makecell{\textbf{Average} \\ \textbf{response} \\ \textbf{time, s} } & \makecell{\textbf{Average} \\ \textbf{quality}} & \makecell{\textbf{\# training} \\ \textbf{sentences}} & \makecell{\textbf{\# control} \\\textbf{sentences}} & \textbf{\# sentences}\\
\midrule
\makecell[l]{\textbf{Stage 1} }   & 1300 & \$2.55/hr & 70 & 80\% & 28 & 179 & 5685 \\
\textbf{Stage 2} & 30 & \$5.42/hr  & 143 & 57\% & 11 & 500 & 2699 \\
\midrule
\bf Human Benchmark & 16 & \$6.3/hr  & 53 & 79\% & 10 & 901 & 2048\\
\bottomrule
\end{tabular}
\caption{Summary of the annotation design details by annotation project.}
\label{tab:annotation_additional}
\end{table*}

\begin{table*}[th!]
\small
\centering
\begin{tabular}{lcccccc}
\toprule
  & \textbf{Acceptable} & \textbf{Morphology} & \textbf{Syntax} & \textbf{Hallucination} & \textbf{Semantics} & \textbf{Average}\\
\midrule
\bf Stage 1 & 0.80 & 0.83 & 0.88 & 0.88 & 0.78 & 0.83\\
\bf Stage 2 & 0.94 & 0.90 & 0.84 & 0.86 & 0.90 & 0.89\\
\midrule
\bf Human Benchmark & 0.85 & 0.81 & 0.82 & 0.88 & 0.80 & 0.85\\
\bottomrule
\end{tabular}
\caption{Per-category WAWA inter-annotator agreement rates by annotation project.}
\label{tab:wawa_granular}
\end{table*}

\subsection{Design Details}
\label{app:design_details}
This subsection summarizes the annotation design details for each annotation project: \textbf{Stage 1: Acceptability Judgments} (Section~\ref{annotation_machine}); \textbf{Stage 2: Violation Categories} (Section~\ref{annotation_machine}); and \textbf{Human Evaluation} (Section~\ref{subsection:human}).

\paragraph{Annotation Project Statistics} \autoref{tab:annotation_additional} describes the following statistics: the number of annotators who participated in the project, the pay rate (\$/hr), the average response time in seconds, the average performance on the control tasks, the number of training and control sentences, and the overall number of sentences.

\paragraph{Inter-annotator Agreement Rates}
\autoref{tab:wawa_granular} presents the per-category IAA rates for each annotation project. The IAA rates are computed with the Worker Agreement with Aggregate (WAWA) coefficient~\cite{ning-etal-2018-joint}. WAWA indicates the average fraction of the annotators' responses that agree with the aggregate answer for each example. The WAWA values are above 0.8 in most cases, which implies a strong agreement between annotators. We observe that the non-expert annotators (\textbf{Stage 1}) have lower average WAVA values than the expert ones (\textbf{Stage 2}; \textbf{Human Evaluation}). Annotators in the \textbf{Human evaluation} project receive high IAA scores on acceptable sentences and sentences containing hallucinations. Although the IAA scores are lower in the other categories, they are still strong.

\newpage
\section{Hyperparameter Values for Baseline Methods}

\begin{table}[h!]
\centering
\small
\begin{tabular}{lcc}
\toprule
\bf Model &\bf  Hyperparameter &\bf Value \\
\midrule
Linear & $\ell_2$ penalty strength & 1 \\
ruGPT-3 & Threshold & -20.92\\ \bottomrule
\end{tabular}
\caption{Optimal hyperparameter values for the linear model and threshold-based baselines.}
\label{tab:hparams_simple}
\end{table}

\begin{table}[h]
\centering
\small
\setlength{\tabcolsep}{4pt}
\begin{tabular}{lccc}
\toprule
\bf Model &\bf  Learning rate &\bf  Weight decay &\bf Batch size \\
\midrule
ruBERT & 3$\cdot 10^{-5}$ & 0.1 & 32 \\
ruRoBERTa & $10^{-5}$ & $10^{-4}$ & 32 \\
ruT5 & $10^{-4}$ & 0 & 128 \\
XLM-R & $10^{-5}$ & 0.1 & 32 \\
RemBERT & $10^{-5}$ & $10^{-4}$ & 64 \\
\bottomrule
\end{tabular}
\caption{Optimal hyperparameter values for finetuned language models.}
\label{tab:hparams_supervised}
\end{table}

\newpage
\clearpage

\begin{figure*}[th]
\section{Acceptability Measure Distribution}
\label{section:app_acceptability_measure}
    \centering
    \includegraphics[width=\linewidth]{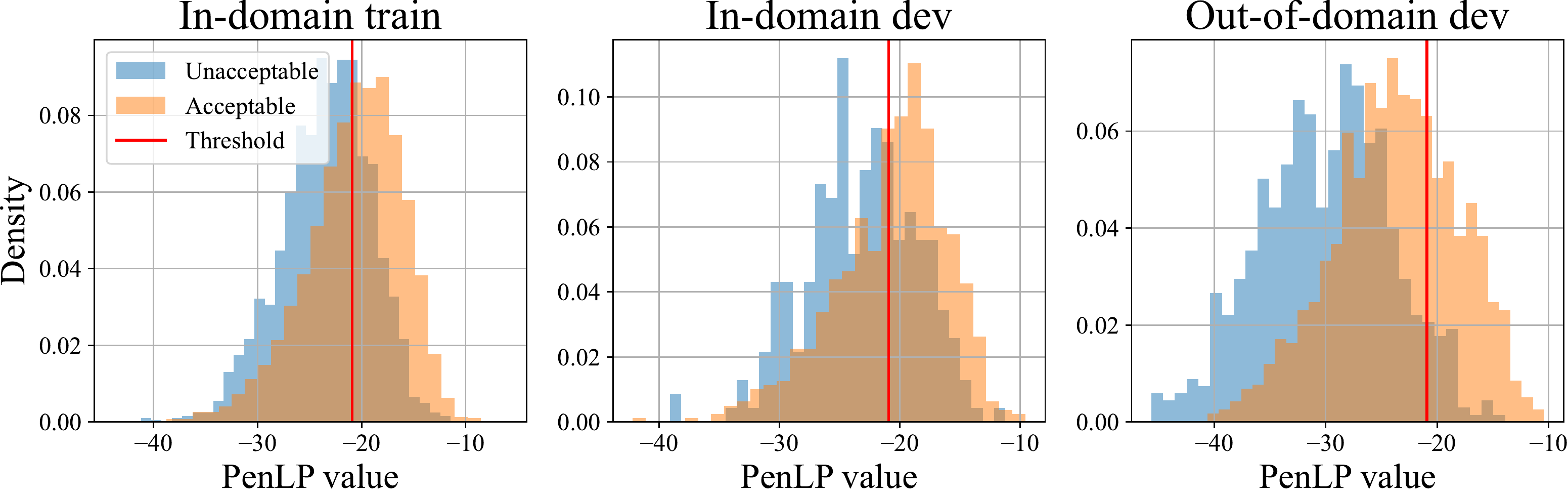}
    \caption{PenLP acceptability measure values for train and validation sets of RuCoLA.}
    \label{fig:penlp}
\end{figure*}

\section{Cross-lingual Evaluation Details} \label{section:appendix_crosslingual}

\begin{table*}[bp]
\setlength{\tabcolsep}{1.5ex}
\vspace{-1ex}
\small
\centering
\begin{tabular}{llccccc}
\toprule
\multirow{2}{*}{\bf Model} &\multicolumn{1}{c}{\bf Training} & \multicolumn{2}{c}{\bf CoLA}  & \multirow{2}{*}{\bf ItaCoLA} & \multicolumn{2}{c}{\bf RuCoLA} \\
 &\multicolumn{1}{c}{\bf data} &\bf In-domain &\bf OOD &    &\bf In-domain &\bf OOD \\
\midrule
\multirow[c]{3}{*}{\bf mBERT} & CoLA &\bf 0.41{\small\ \pm\ 0.04} &\bf 0.35{\small\ \pm\ 0.03} & 0.03{\small\ \pm\ 0.03} & -0.04{\small\ \pm\ 0.02} & 0.07{\small\ \pm\ 0.02} \\

 & ItaCoLA & 0.07{\small\ \pm\ 0.05} & 0.09{\small\ \pm\ 0.05}  &\bf 0.36{\small\ \pm\ 0.04} & 0.0{\small\ \pm\ 0.03} & 0.0{\small\ \pm\ 0.02} \\
 & RuCoLA & -0.02{\small\ \pm\ 0.04} & 0.01{\small\ \pm\ 0.05}  & 0.0{\small\ \pm\ 0.02} &\bf 0.18{\small\ \pm\ 0.02} &\bf 0.15{\small\ \pm\ 0.03} \\
 \midrule
\multirow[c]{3}{*}{\bf XLM-R\textsubscript{Base}} & CoLA &\bf 0.55{\small\ \pm\ 0.03} &\bf 0.51{\small\ \pm\ 0.02}  & 0.2{\small\ \pm\ 0.02} & 0.1{\small\ \pm\ 0.01} & 0.3{\small\ \pm\ 0.02} \\

 & ItaCoLA & 0.05{\small\ \pm\ 0.07} & 0.03{\small\ \pm\ 0.06}  &\bf 0.25{\small\ \pm\ 0.08} & 0.01{\small\ \pm\ 0.02} & 0.01{\small\ \pm\ 0.02} \\
 & RuCoLA & 0.05{\small\ \pm\ 0.08} & 0.06{\small\ \pm\ 0.06}  & 0.01{\small\ \pm\ 0.05} &\bf 0.23{\small\ \pm\ 0.05} & \bf 0.2{\small\ \pm\ 0.06} \\
\midrule
\multirow[c]{3}{*}{\bf XLM-R} & CoLA &\bf 0.61{\small\ \pm\ 0.02} &\bf 0.57{\small\ \pm\ 0.03}  & 0.3{\small\ \pm\ 0.02} & 0.2{\small\ \pm\ 0.03} &\bf 0.42{\small\ \pm\ 0.02} \\

 & ItaCoLA & 0.3{\small\ \pm\ 0.03} & 0.32{\small\ \pm\ 0.03}  &\bf 0.52{\small\ \pm\ 0.03} & 0.13{\small\ \pm\ 0.03} & 0.24{\small\ \pm\ 0.02} \\
 & RuCoLA & 0.24{\small\ \pm\ 0.12} & 0.26{\small\ \pm\ 0.13} & 0.17{\small\ \pm\ 0.08} &\bf 0.36{\small\ \pm\ 0.05} & 0.4{\small\ \pm\ 0.06} \\
\midrule
\multirow[c]{3}{*}{\bf RemBERT} & CoLA &\bf 0.65{\small\ \pm\ 0.02} &\bf 0.6{\small\ \pm\ 0.02}  & 0.29{\small\ \pm\ 0.03} & 0.17{\small\ \pm\ 0.02} &\bf 0.44{\small\ \pm\ 0.03} \\

 & ItaCoLA & 0.48{\small\ \pm\ 0.04} & 0.44{\small\ \pm\ 0.04}  &\bf 0.52{\small\ \pm\ 0.02} & 0.15{\small\ \pm\ 0.03} & 0.39{\small\ \pm\ 0.04} \\
 & RuCoLA & 0.46{\small\ \pm\ 0.03} & 0.44{\small\ \pm\ 0.01}  & 0.29{\small\ \pm\ 0.02} &\bf 0.41{\small\ \pm\ 0.02} & \bf0.44{\small\ \pm\ 0.02} \\
\bottomrule
\end{tabular}
\caption{MCC for cross-lingual acceptabiliy classification. The best score is in bold.}
\label{tab:crosslingual}
\end{table*}

Here, we describe the setup of experiments outlined in Section~\ref{sect:crosslingual}. We use four models: Multilingual BERT-base-cased (110M parameters), XLM-RoBERTa-base (or XLM-R\textsubscript{Base}, 270M parameters), XLM-RoBERTa-large (or XLM-R, 550M parameters), and RemBERT. For each pair of \textbf{source} and \textbf{target} languages, we train and tune the hyperparameters on the train and development sets of the \textbf{source} language respectively and compute the final metrics on the \textbf{target} language. We also include the pairs consisting of the same language (i.e., the same dataset) for source and target to provide an upper bound for classification quality. We do not report accuracy both for brevity and because the test set leaderboard for CoLA reports only the MCC values.

The results are averaged over ten different random seeds: we use mean MCC on the development set for hyperparameter tuning and report the average and the standard deviation of the test set metrics. We optimize the validation score with grid search with respect to the following hyperparameters: learning rate (the search space is $\{10^{-5},\ 3\cdot 10^{-5},\ 5\cdot 10^{-5}\}$), batch size $\{32,\ 64\}$, weight decay $\{10^{-4},\ 10^{-2},\ 0.1\}$. Each model is trained for 5 epochs with early stopping based on the validation MCC.

\autoref{tab:crosslingual} shows the results 
of our study. The RemBERT model outperforms other cross-lingual encoders, which aligns with the results of~\citet{chung2020rethinking}.
A key observation is that although the quality on the out-of-domain data is indeed similar for all source languages, for the in-domain test set of RuCoLA, the cross-lingual generalization gap remains quite large, similarly to other studied datasets.
This indicates that source on other datasets does not induce the language understanding capabilities that are necessary for estimating acceptability in the Russian language, which is expected given its typological differences from the English and Italian languages.

\end{document}